\documentclass{article}
\usepackage{ijcai20}

\usepackage{times}

\usepackage{soul}
\usepackage{url}
\usepackage[hidelinks]{hyperref}
\usepackage[utf8]{inputenc}
\usepackage[small]{caption}
\usepackage{graphicx}
\usepackage{amsmath}
\usepackage{amsthm}
\usepackage{booktabs}
\usepackage{latexsym} 

\usepackage{subcaption}
\usepackage{amssymb}
\usepackage{algorithm}
\usepackage{algpseudocode}
\usepackage[dvipsnames]{xcolor}
\definecolor{Dgreen}{rgb}{0.0, 0.5, 0.0}
\newcommand{\devnull}[1]{}

\newcommand{\ouralgo}{PB-MHB}
\newcommand{\ouralgolong}{Position Based Metropolis-Hastings Bandit}

\input{macros_math}

\title{Position-Based Multiple-Play Bandits\\ with Thompson Sampling\footnote{A shorter version of this paper has been accepted at IDA 2021 under the title: Bandit Algorithm for Both Unknown Best Position and Best Item Display on Web Pages.}}

\author{
Camille-Sovanneary Gauthier$^{1,2}$\and
Romaric Gaudel$^{3}$\And
Elisa Fromont$^{2,4}$\\
\affiliations
$^1$Louis Vuitton\\
$^2$Université de Rennes 1, IRISA/INRIA rba, $^4$IUF\\
$^3$Univ. Rennes, Ensai, CNRS, CREST - UMR 9194\\
\emails
camille-sovanneary.gauthier@louisvuitton.com,
romaric.gaudel@ensai.fr,
elisa.fromont@irisa.fr
}

\begin{document}

\maketitle

\begin{abstract}
Multiple-play bandits aim at displaying relevant items at relevant positions on a web page.
    We introduce a new bandit-based algorithm, \ouralgo{}, for online recommender systems which uses the Thompson sampling framework. This algorithm handles a display setting governed by the position-based model.
    Our sampling method does not require as input the probability of a user to look at a given position in the web page which is, in practice, very difficult to obtain. Experiments on simulated and real datasets show that our method, with fewer prior information, deliver better recommendations than state-of-the-art algorithms.
\end{abstract}

    \section{Introduction}
    
    Online recommender systems choose a good item to recommend to a user among a list of $N$ potential items. The relevance of the item is measured by the users' indirect feedback: clicks, time spent looking at the item, rating, purchases etc. Since feedback is only available when an item is presented to a user, recommender systems need to present both attractive items (a.k.a. \emph{exploit}) to please the current user, and some items with an uncertain relevance (a.k.a. \emph{explore}) to reduce this uncertainty and do better recommendations to future users. They face the \emph{exploration-exploitation dilemma} expressed by the \emph{multi-armed bandit setting} \cite{Auer2002}.
    
    On websites, online recommender systems select $L$ items per time-stamp, corresponding to $L$ specific positions in which to display an item. Typical examples of such systems are (i) a list of news, visible one by one by scrolling; (ii) a list of products, arranged by rows; or (iii) advertisements spread in a web page. To be selected (clicked) by a user in such context, an item needs to be relevant by itself, but also to be displayed at the right position. Several models express the way a user behaves while facing such a list of items \cite{Richardson2007,Craswell2008} and they have been transposed to the bandit framework \cite{Kveton2015a,Komiyama2017}.
    
    In this paper, we will focus on the \emph{Position-Based Model} (PBM) \cite{Richardson2007}. This model assumes that the probability to click on an item $i$ in position $\ell$ results only from the combined impact of this item and its position: items displayed at other positions do not impact the probability to consider the item at position $\ell$. This assumption is relevant when items are spread in a web page or displayed on several rows all at once. PBM also gives a user the opportunity to give more than one feedback: she may click on all the items relevant for her. It means we are facing the so-called \emph{multiple-play} semi-bandit setting \cite{Chen2013}. This setting is particularly interesting when the display is dynamic, as often on modern web pages, and may depend on the reading direction of the user (which varies from one country to another) and on the ever-changing layout of the page.  
    
    \paragraph{Contributions} We introduce \ouralgo{} (\ouralgolong{}), a bandit algorithm designed to handle PBM with a Thompson sampling framework. \ouralgo{} improvement w.r.t. previous attempts in this research line \cite{Komiyama2015,Lagree2016} is two-fold. First, it  does not require the knowledge of the probability of a user to look at a given position: it learns this probability from past recommendations/feedbacks. Secondly, \ouralgo{} is useful even when the click-probabilities are extreme, namely close to 0 or to 1. We believe that this is a major advantage in many commercial applications where the click-probabilities are closer to 0.1 or 0.01 than to 0.5. Both improvements result from the use of the Metropolis-Hastings framework to sample parameters given their \emph{a posteriori} distribution. 
    
    The paper is organized as follows: Section \ref{sec:related} presents the related work and Section \ref{sec:setting} precisely defines our target setting. \ouralgo{} is introduced in Section \ref{sec:algorithm} and is experimentally compared to state-of-the-art algorithms in Section \ref{sec:exp}. We conclude in Section \ref{sec:conclu}.

    
    \section{Related work}\label{sec:related}
    The Position-Based Model (PBM) \cite{Richardson2007,Craswell2008} relies on two vectors of parameters: $\thetav \in [0,1]^N$ and $\kappav \in [0,1]^L$, where $\thetav_i$ is the probability for the user to click on item $i$ when she observes that item, and $\kappav_\ell$ is the probability for the user to observe the position $\ell$. These parameters are unknown, but they may be inferred from user behavior data: we need to first record the user feedback (click vs. no-click per position) for each set of displayed items, then we may apply an \emph{expectation-maximization} framework to compute the maximum a posteriori values for $(\thetav, \kappav)$ given these data \cite{Chuklin2015}. 
    
    PBM is transposed to the bandit framework in \cite{Komiyama2015,Lagree2016,Komiyama2017}. 
    \cite{Komiyama2015} and \cite{Lagree2016} propose two approaches based on a Thompson sampling framework, with two different sampling strategies. \cite{Lagree2016} also introduces several approaches based on the \emph{optimism in face of uncertainty} principle \cite{Auer2002}. However, the approaches in \cite{Komiyama2015,Lagree2016} assume $\kappav$ known beforehand. \cite{Komiyama2017} proposes the only approach learning both $\thetav$ and $\kappav$ while recommending but it is not based, as ours, on Thompson sampling.
    
    More recently,  \cite{Katariya2017a,Katariya2017b} analyzed a slightly different framework: the \emph{rank-1 bandit}. In this framework, the probability for an item $i$ to be clicked in position $\ell$ is also $\thetav_i\kappav_\ell$. The difference lies in the interaction setting. While in our setting the algorithm has to choose an item per position at each time-stamp, in the rank-1 bandit setting, the algorithm only picks one item and the position where it will be displayed.
   We may also mention \cite{Kawale2015} which has to choose an item per time-stamp, the user (counterpart of the position in our setting) is considered fixed by an external process. Nonetheless, in this setting, the representation of each item / user is richer (a vector as opposed to a value in the previously cited settings) and the feedback is a rate (usually in the range [0,5]) while we restrict ourselves to binary feedback (not clicked vs. clicked).

    The \emph{cascading model} \cite{Craswell2008} is another popular user's behavior model. It assumes that the positions are observed in a known order and that the user leaves the website as soon as she clicks on an item\footnote{Some refined models assume a probability to leave. With these models, the user may click on several items.}. More specifically, if the user clicks on the item in position $\ell$, she will not look at the following positions: $\ell+1,\dots,L $. This setting has been extensively studied within the bandit framework \cite{Zong2016,Katariya2016,Kveton2015a,Kveton2015b,Li2016,Combes2015,Cheung2019}. Still, the assumption of cascading models regarding the order of observation is irrelevant when considering items spread in a page or items arranged by rows.

    \section{Recommendation setting}
    \label{sec:setting}
    The proposed approach handles the following online recommendations setting: at each time-stamp $t$, the recommender system chooses $L$ ordered distinct items $\iv(t) = \left(\iv_\ell(t)\right)_{\ell=1}^L$ among a set of $N$ items. The user observes each position $\ell$ with a probability $\kappav_\ell$, and if the position is observed, the user clicks on the item $\iv_\ell(t)$ with a probability $\thetav_{\iv_\ell(t)}$. We denote $\rv_\ell(t)$ the reward obtained at time-stamp $t$ in position $\ell$, namely $1$ if the user did observe the position $\ell$ and clicked on item $\iv_\ell(t)$, and $0$ otherwise. We assume that each draw is independent, meaning 
    \begin{equation}\label{eq:PBM}
        \rv_\ell(t) \mid \iv_\ell(t) \stackrel{iid.}{\thicksim} \Ber\left(\thetav_{\iv_\ell(t)}\kappav_\ell\right),
    \end{equation}
    where $\Ber$ is the Bernoulli distribution, or in other words
    \begin{equation*}
        \begin{cases}
        \PP\left(\rv_\ell(t) = 1 \mid \iv_\ell(t)\right) = \thetav_{\iv_\ell(t)}\kappav_\ell,\\
        \PP\left(\rv_\ell(t) = 0 \mid \iv_\ell(t)\right) = 1 - \thetav_{\iv_\ell(t)}\kappav_\ell.
        \end{cases}
    \end{equation*}
    
    The recommender system aims at maximizing the \emph{cumulative reward}, namely the total number of clicks gathered from time-stamp 1 to time-stamp $T$: $\sum_{t=1}^T\sum_{\ell=1}^L\rv_\ell(t)$. 
     Without loss of generality, let's assume that $\max_\ell \kappav_\ell = 1$. To keep the notations simple, let's also assume that $\thetav_1 > \thetav_2 > \dots > \thetav_N$, and $\kappav_1=1 > \kappav_2 > \dots > \kappav_L$.\footnote{The algorithms and the experiments only assume $\kappav_1=1$.} The best recommendation is then $\iv^* =  \left(1, 2, \dots, L\right)$,  which leads to the expected  instantaneous reward $\mu^* = \sum_{\ell=1}^L\thetav_\ell\kappav_\ell$.
    
    The parameters $\thetav$ and $\kappav$ are unknown from the recommender system. It has to infer the best recommendation from the recommendations and the rewards gathered at previous time-stamps, denoted $D(t) = \{\iv(1),\dots, \iv(t-1), \rv(1),\dots, \rv(t-1)\}$. This corresponds to the \emph{bandit setting} where it is usual to consider the \emph{(cumulative pseudo-)regret}
    \begin{align}
    R_T &\stackrel{def}{=} \sum_{t=1}^T\sum_{\ell=1}^L \EE\left[\rv_\ell(t) \mid \iv^*_\ell\right]- \sum_{t=1}^T\sum_{\ell=1}^L \EE\left[\rv_\ell(t) \mid \iv_\ell(t)\right]\\
     &= \mu^*T - \sum_{t=1}^T\sum_{\ell=1}^L \thetav_{\iv_\ell(t)}\kappav_\ell.
     \label{PseudoRegCum}
    \end{align} 
    The regret $R_T$ denotes the cumulative expected loss of the recommender system w.r.t. the oracle recommending the best items at each time-stamp. Hereafter we aim at an algorithm which minimizes the expectation of $R_T$ w.r.t. its choices.
    
 \begin{algorithm}[t]
     \caption{\ouralgo{}, Metropolis-Hastings based bandit for Position-Based Model}\label{alg:\ouralgo{}}
     \begin{algorithmic}
         \For{$t = 1,\dots$}
         \State draw $(\tilde\thetav, \tilde{\kappav}) \sim \PP\left(\thetav, \kappav | D(t)\right)$ 
         using Algorithm \ref{alg:MH}
         \State display the $L$ items with greatest value in $\tilde\thetav$, ordered by decreasing values of  $\tilde\kappav$
         \State get rewards $\rv(t)$
         \EndFor
     \end{algorithmic}
 \end{algorithm}
 
 \section{\ouralgo{} algorithm}\label{sec:algorithm}

    We handle the setting presented in the previous section with the online recommender system depicted by Algorithm \ref{alg:\ouralgo{}} and referred to as \ouralgo{} (for \ouralgolong{}). This algorithm is based on the Thompson sampling framework \cite{Thompson1933,Agrawal2017}. First, we look at rewards with a fully Bayesian point of view: we assume that they follow the statistical model depicted in section \ref{sec:setting}, and we choose a uniform prior on the parameters $\thetav$ and $\kappav$. Therefore the posterior probability for these parameters given the previous observations $D(t)$ is
    \begin{equation}\label{eq:posterior}
        \PP\left(\thetav, \kappav | D(t)\right)
    \propto \prod_{i=1}^{N}\prod_{\ell=1}^{L}\left(\thetav_i\kappav_\ell\right)^{S_{i,\ell}(t)}\left(1-\thetav_i\kappav_\ell\right)^{F_{i,\ell}(t)},
    \end{equation}%
    where $S_{i,\ell}(t)=\sum_{s=1}^{t-1}\ind_{\iv_\ell(s)=i}\ind_{\rv_\ell(s)=1}$ denotes the number of times the item $i$ has been clicked while being displayed in position $\ell$ from time-stamp $1$ to $t-1$, and $F_{i,\ell}(t)=\sum_{s=1}^{t-1}\ind_{\iv_\ell(s)=i}\ind_{\rv_\ell(s)=0}$ denotes the number of times the item $i$ has not been clicked while being displayed in position $\ell$ from time-stamp $1$ to $t-1$.
    
    Second, we choose the recommendation $\iv(t)$ at time-stamp $t$ according to its posterior probability of being the best arm. To do so, we denote $(\tilde\thetav, \tilde{\kappav})$ a sample of parameters $(\thetav, \kappav)$ according to their posterior probability, we keep the best items given $\tilde\thetav$, and we display them in the right order given $\tilde{\kappav}$.
    
    \algloopdefx[WithProbability]{WithProbability}[1]{\textbf{with prob.} #1}
    \begin{algorithm}[t]
        \caption{Metropolis-Hastings applied to the distribution of Equation \eqref{eq:posterior}}\label{alg:MH}
        \begin{algorithmic}[1]
            \Require{$D(t)$: previous recommendations and rewards}
            \Require{$\sigma = {c}/{\sqrt{t}}$: Gaussian random-walk steps width}
            \Require{$m$: number of iterations}
            \State draw $(\thetav, \kappav)$ after uniform distribution \label{algline:initial}
            \State $\kappav_1 \gets 1$
            \For{$s = 1,\dots, m$}
            \For{$i = 1,\dots, N$}
            \Repeat\label{algline:itemibegin}\label{algline:proposalbegin}
            \State draw $\tilde\theta \sim \Norm\left(\thetav_i, \sigma\right)$
            \Until{$0\leqslant\tilde\theta\leqslant1$}\label{algline:proposalend}
            \WithProbability{$
            \min\left(1,
            \frac{\PP_i\left(\tilde\theta | \thetav_{-i}, \kappav, D(t)\right)}{\PP_i\left(\thetav_i | \thetav_{-i}, \kappav, D(t)\right)}
            \frac{\Delta\Phi_\sigma\left(\thetav_i\right)}{\Delta\Phi_\sigma\left(\tilde\theta\right)}
            \right)
            $}\label{algline:accept}
            \State $\thetav_i \gets \tilde\theta$\label{algline:itemiend}
            \EndFor
            \For{$\ell = 2,\dots, L$}
            \Repeat
            \State draw $\tilde\kappa\sim \Norm\left(\kappav_\ell, \sigma\right)$
            \Until{$0\leqslant\tilde\kappa\leqslant1$}
            \WithProbability{$\min\left(1, \frac
            {\PP_\ell\left(\tilde\kappa | \thetav, \kappav_{-\ell}, D(t)\right)}
            {\PP_\ell\left(\kappav_\ell | \thetav, \kappav_{-\ell}, D(t)\right)}
            \frac{\Delta\Phi_\sigma\left(\kappav_\ell\right)}{\Delta\Phi_\sigma\left(\tilde\kappa\right)}
            \right)$}
            \State $\kappav_\ell \gets \tilde\kappa$
            \EndFor
            \EndFor
            \State \Return $(\thetav, \kappav)$
        \end{algorithmic}
    \end{algorithm}
    
    \subsection{Sampling w.r.t. the posterior distribution}
    The posterior probability \eqref{eq:posterior} does not correspond to a well-known distribution. We handle it thanks to a carefully designed Metropolis-Hastings algorithm \cite{Neal1993} (cf. Algorithm \ref{alg:MH}). This algorithm consists in building a sequence of $m$ samples $(\thetav^{(1)}, \kappav^{(1)}), \dots, (\thetav^{(m)}, \kappav^{(m)})$ such that $(\thetav^{(m)}, \kappav^{(m)})$ follows a good approximation of the targeted distribution. It is based on a Markov chain on parameters $(\thetav, \kappav)$ which admits the targeted probability distribution as its unique stationary distribution.
    
    At iteration $s$, the sample $(\thetav^{(s)}, \kappav^{(s)})$ moves toward sample $(\thetav^{(s+1)}, \kappav^{(s+1)})$ by applying $(N+L-1)$ transitions: one per item and one per position except for $\kappa_{1}$. Let's start by focusing on the transition regarding item $i$ (Lines \ref{algline:itemibegin}--\ref{algline:itemiend}) and denote $(\thetav, \kappav)$ the sample before the transition.
    
    The algorithm aims at sampling a new value for $\thetav_i$ according to its posterior probability given other parameters and the previous observations $D(t)$:
    \begin{equation}\label{eq:thetai}
    \PP_i\left(\thetav_i | \thetav_{-i}, \kappav, D(t)\right) \propto \prod_{\ell=1}^{L}{\thetav_i}^{S_{i,\ell}(t)}\left(1-\thetav_i\kappav_\ell\right)^{F_{i,\ell}(t)},  
    \end{equation}
    where  $\thetav_{-i}$ denotes the components of $\thetav$ except for the $i$-th one.
    This transition consists in two steps:
    \begin{enumerate}
        \item draw a \emph{candidate} value $\tilde \theta$ after a \emph{proposal} probability distribution $q\left(\tilde\theta \mid \thetav_i, \thetav_{-i}, \kappav, D(t) \right)$ to be discussed later on;
        \item \emph{accept} that candidate or keep the previous sample: $\thetav_i$ gets $\tilde\theta$ with probability
        $\min\left(1,
        \frac{\PP_i\left(\tilde\theta | \thetav_{-i}, \kappav, D(t)\right)}
        {\PP_i\left(\thetav_i | \thetav_{-i}, \kappav, D(t)\right)}
        \frac{q\left(\thetav_i \mid\tilde\theta, \thetav_{-i}, \kappav, D(t) \right)}
        {q\left(\tilde\theta \mid \thetav_i, \thetav_{-i}, \kappav, D(t) \right)}
        \right)$, and gets $\thetav_i$ otherwise.
    \end{enumerate}

    This acceptance step yields two behaviours: 
    \begin{itemize}
        \item  $\frac{\PP_i\left(\tilde\theta | \thetav_{-i}, \kappav, D(t)\right)}
        {\PP_i\left(\thetav_i | \thetav_{-i}, \kappav, D(t)\right)}$ measures how likely the candidate value is compared to the previous one, w.r.t. the posterior distribution,
        \item $\frac{q\left(\thetav_i \mid\tilde\theta, \thetav_{-i}, \kappav, D(t) \right)}
        {q\left(\tilde\theta \mid \thetav_i, \thetav_{-i}, \kappav, D(t) \right)}$ prevents preferring candidates easily reached by the proposal $q$.  
    \end{itemize}

    

    Algorithm \ref{alg:MH} uses a truncated Gaussian random-walk proposal for the parameter $\thetav_i$, with a Gaussian step of \emph{standard deviation} $\sigma$ (see Lines \ref{algline:proposalbegin}--\ref{algline:proposalend}). Note that due to the truncation, the probability to get the proposal $\tilde\theta$ starting from $\thetav_i$ is
    $$q\left(\tilde\theta \mid \thetav_i, \thetav_{-i}, \kappav, D(t) \right)
    =
    \phi(\tilde\theta\mid \thetav_i, \sigma)
    /\Delta\Phi_\sigma(\thetav_i),$$
    where $\phi(\cdot\mid \thetav_i, \sigma)$ is the probability associated to the Gaussian distribution with mean $\thetav_i$ and standard deviation $\sigma$, $\Phi(\cdot\mid \thetav_i, \sigma)$ is its cumulative distribution function, and $\Delta\Phi_\sigma(\thetav_i) = \Phi(1\mid \thetav_i, \sigma) - \Phi(0\mid \thetav_i, \sigma)$.
    The probability to get the proposal $\thetav_i$ starting from $\tilde\theta$ is similar, which reduces the ratio of proposal probabilities at Line \ref{algline:accept} to
    $$
    \frac{q\left(\thetav_i \mid\tilde\theta, \thetav_{-i}, \kappav, D(t) \right)}
        {q\left(\tilde\theta \mid \thetav_i, \thetav_{-i}, \kappav, D(t) \right)}
    =
    \frac{\Delta\Phi_\sigma\left(\thetav_i\right)}{\Delta\Phi_\sigma\left(\tilde\theta\right)}.
    $$ 
    
    The transition regarding parameter $\kappav_\ell$ involves the same framework: the proposal is a truncated Gaussian random-walk step and aims at the probability
        \begin{equation}\label{eq:kappal}
    \PP_\ell\left(\kappav_\ell | \thetav, \kappav_{-\ell}, D(t)\right) \propto \prod_{i=1}^{N}{\kappav_\ell}^{S_{i,\ell}(t)}\left(1-\thetav_i\kappav_\ell\right)^{F_{i,\ell}(t)}.  
    \end{equation}

    \subsection{Overall complexity}
    The computational complexity of Algorithm \ref{alg:\ouralgo{}} is driven by the number of random-walk steps done per recommendation: $m(N+L-1)$, which is controlled by the parameter $m$. This parameter corresponds to the burning period: the number of iterations required by the Metropolis-Hastings algorithm to draw a point $(\thetav^{(m)}, \kappav^{(m)})$ almost independent from the initial one.  As we demonstrate in the following experiments, the required value for $m$ remains reasonable in our context. We drastically reduce $m$ by starting the Metropolis-Hasting call from the point used to recommend at previous time-stamp: this corresponds to replacing Line \ref{algline:initial} in Algorithm \ref{alg:MH} by:\\\ref{algline:initial}: $(\thetav, \kappav) \gets (\tilde\thetav, \tilde\kappav)$ used for the previous recommendation.

    
    \section{Experiments}\label{sec:exp}
    In this section we demonstrate the benefit of the proposed approach both on artificial and real-life datasets.
    
    \subsection{Datasets}
    In the experiments, the online recommender systems are required to deliver $T$ consecutive recommendations, their feedbacks being drawn from a PBM distribution (Equation \eqref{eq:PBM}). We consider two settings denoted \emph{purely simulated} and \emph{behavioral} in the remaining. With the \emph{purely simulated} setting, we choose the value of the parameters $(\thetav, \kappav)$ to highlight the properties of the proposed approach. Namely, we consider $N=10$ items, $L=5$ positions, and $\kappav = [1,0.75,0.6,0.3,0.1]$. The range of values for $\thetav$ is either close to zero ($\kappav = [0.15, 0.1, 0.1, 0.05, 0.05, 0.01, \dots, 0.01]$), or close to one ($\kappav = [0.99, 0.95, 0.9, 0.85, 0.8, 0.75, \dots, 0.75]$), or characteristic of the one encountered for website interactions ($\thetav = [0.3, 0.2, 0.15, 0.15, 0.15, 0.10,0.05,0.05,0.01,0.01]$).
    
    With the \emph{behavioral} setting, the values for $\kappav$ and $\thetav$ are obtained from users behavior as in \cite{Lagree2016}. More specifically, $\kappav$ and $\thetav$ are extracted from real KDD Cup 2012 track 2 dataset, which consists of session logs of soso.com, a Tencent's search engine. It tracks clicks and displays of advertisement on a search engine result web-page, w.r.t. the user query. Each of the 150M lines contains information about the search (UserId, QueryId\dots) and the ads displayed (AdId, Position, Click, Impression). We are looking for the best ads per query, namely the ones with a higher probability to be clicked.
    
    To follow the previous work, instead of looking for the probability to be clicked per display, we target the probability to be clicked per session. This amounts to discarding the information \emph{Impression}.  
    We also filter the logs to restrict the analysis to (query, ad) couples with enough information: for each query, ads are excluded if they were displayed less than 1,000 times at any of the 3 possible positions. Then, we filter queries that have less than 5 ads satisfying the previous condition. We end up with 8 queries and from 5 to 11 ads per query.
    Finally, for each query $q$, the parameters $(\thetav^{[q]},\kappav^{[q]})$ are set from the \emph{Singular Value Decomposition} (SVD) of the matrix $\Mm^{[q]} \in \mathbb{R}^{N\times L}$ which contains the probability to be clicked for each item in each position. By denoting $\zeta^{[q]}$, the greatest singular value of $\Mm^{[q]}$, and $\uv^{[q]}$ (respectively $\vv^{[q]}$) the left (resp. right) singular vector associated to $\zeta^{[q]}$, we set
    \begin{align*}
    \thetav^{[q]} &\stackrel{def}{=}\vv^{[q]}_1\zeta^{[q]}\uv^{[q]}&
    \kappav^{[q]} &\stackrel{def}{=}\vv^{[q]} / \vv^{[q]}_1,
    \end{align*}
    such that $\kappav^{[q]}_1=1$, and ${\thetav^{[q]}}^T\kappav^{[q]} = \zeta{\uv^{[q]}}^T\vv^{[q]}$.  Table \ref{tab:data_stats} gives more insight regarding the parameters associated to each query.
    
    \begin{table}
    \centering
    \caption{Number of ads and range of values for the parameters $\thetav$ and $\kappav$ inferred from the KDD Cup 2012 track 2 dataset.}
    \label{tab:data_stats}
    \begin{tabular}{cccccc}
    \toprule
        \textbf{$N$ (\#ads)} & \textbf{$\min_i \thetav_i$} & \textbf{$\max_i \thetav_i$} & \textbf{$\kappav_1$} & \textbf{$\kappav_2$} & \textbf{$\kappav_3$}\\
        \midrule
5&0.016&0.077&1.00&0.503&0.403\\ 
5&0.031&0.050&1.00&0.486&0.330\\ 
6&0.025&0.067&1.00&0.491&0.345\\ 
6&0.017&0.069&1.00&0.546&0.529\\ 
6&0.004&0.148&1.00&0.411&0.275\\ 
8&0.108&0.146&1.00&0.178&0.101\\ 
11&0.022&0.149&1.00&0.473&0.328\\ 
11&0.022&0.084&1.00&0.478&0.349\\ 
\bottomrule
    \end{tabular}
    \end{table}

    Note that we use SVD to set the parameters $(\thetav^{[q]},\kappav^{[q]})$, while a more appropriate but slower inference method would be \emph{expectation-maximization} \cite{Chuklin2015}. Still, both inference processes lead to similar values, as can be seen by comparing Table \ref{tab:data_stats} and Table 1 in \cite{Lagree2016}.

    \begin{figure*}[t]%
    \centering%
    \begin{subfigure}[b]{0.325\linewidth}
        \centering
    \includegraphics[width=\linewidth]{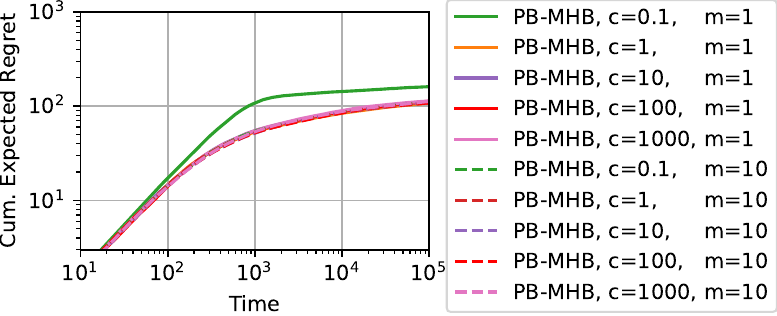}
        \caption{$\thetav$ close to 0}
        \label{fig:cpas_small}
    \end{subfigure}%
    \hfill%
    \begin{subfigure}[b]{0.325\linewidth}
        \centering
    \includegraphics[width=\linewidth]{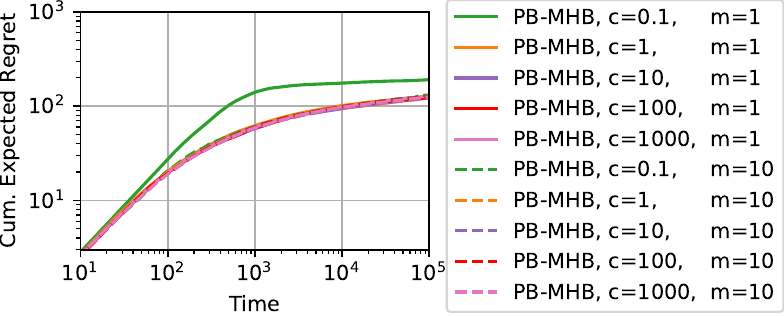}
        \caption{$\thetav$ close to real data}
        \label{fig:cpas_std}
    \end{subfigure}%
    \hfill%
    \begin{subfigure}[b]{0.325\linewidth}
        \centering
    \includegraphics[width=\linewidth]{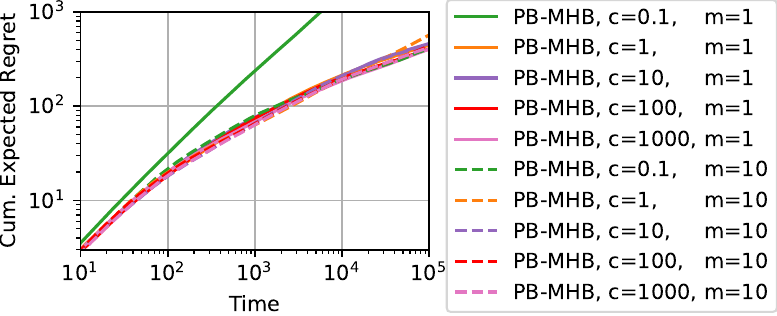}
        \caption{$\thetav$ close to 1}
        \label{fig:cpas_big}
    \end{subfigure}%
    \caption{Cumulative regret w.r.t. time on purely simulated data. Impact of the width $c/\sqrt{t}$ of Gaussian random-walk steps, and the number $m$ of Metropolis-Hastings iterations per recommendation.}
    \label{cpas}
    \end{figure*}  

    \subsection{Competitors}
    We compare the performance of \ouralgo{} with the performance of BC-MPTS \cite{Komiyama2015}, PBM-TS \cite{Lagree2016},  $\varepsilon_n$-greedy \cite{Auer2002}, and Greedy algorithms. BC-MPTS and  PBM-TS assume that $\kappav$ is known beforehand. Therefore, we have considered two versions of these algorithms: one, unrealistic, using the "real" $\kappav$ as input -- \emph{semi-oracle} version, the other using SVD at each time-stamp to infer $\kappav$ from the collected data -- \emph{greedy} version.
    Both algorithms are also based on the Thompson sampling framework. BC-MPTS draws each component of $\thetav$ according to: 
    \begin{equation}\label{eq:BetaBCMPTS}
    \thetav_{i}(t) \sim \Beta\left(S_{i}(t)+1, N^{pseudo}_{i}(t) - S_{i}(t)+1\right),
    \end{equation}
    with $S_{i}(t) = \sum_{\ell=1}^LS_{i,\ell}(t)$ the sum of the clicks obtained by item $i$ over all the positions until time-stamp $t-1$, $N^{pseudo}_{i}(t)= \sum_{\ell=1}^L\kappav_\ell^{(t)}\sum_{s=1}^{t-1}\ind_{\iv_\ell(s)=i}$ the pseudo-expected number of times item $i$ has been observed, and $\Beta$ the \emph{beta distribution}. Thus, BC-MPTS draws the parameter $\thetav$ according to an approximation of its posterior distribution, whereas we sample our parameters according to their exact posterior distribution thanks to the Metropolis-Hastings sampling. 
    
    PBM-TS, also starts by sampling $\thetav$ according to an approximation, but corrects it thanks to rejection sampling \cite{vonNeumann1951}. The target distribution for $\thetav_i$ is given by Equation \eqref{eq:thetai} and the proposal distribution is
    \begin{equation}\label{eq:PropPBMTS}
    \thetav_{i}(t) \sim \Beta(S_{i,\ell_{max}}(t)+1,F_{i,\ell_{max}}(t)+1)/\kappa_{\ell_{max}}^{(t)},
    \end{equation}
    where $\ell_{max}=\argmax_{1\leqslant \ell \leqslant L}(S_{i,\ell}(t)+F_{i,\ell}(t))$.

    Finally, we compare \ouralgo{} to a Greedy algorithm, and its $\epsilon_n$-Greedy version.
    At each time-stamp $t$, the parameters  $(\thetav, \kappav)$ are set applying the SVD to the collected data.
    Let's denote $\hat{\iv}(t)$ the recommendation with the highest expected reward given the inferred values $(\thetav, \kappav)$. The Greedy algorithm consists in recommending $\hat{\iv}(t)$. Since this algorithm never explores, it may end-up recommending a sub-optimal affectation. $\epsilon_n$-Greedy counters this by randomly replacing each item of the recommendation with a probability $\varepsilon(t) = c / t$, where $c$ is a hyper-parameter to be tuned. In the following, we plot the results obtained with the best possible value for $c$, while trying $c$ in $\{10^0, 10^1, \dots, 10^6\}$.
    
    Note that Greedy, $\epsilon_n$-Greedy, and greedy versions of BC-MPTS and PBM-TS use SVD to infer $\thetav$ and $\kappav$. Again, a much proper inference process would be to pick the \emph{Maximum A Posteriori} (MAP) value for $(\thetav, \kappav)$ given the collected data. This MAP value may be inferred with an expectation-maximisation (EM) approach \cite{Chuklin2015}. We used EM with the Pyclick implementation \cite{Pyclick} for both Greedy and $\epsilon_n$-Greedy algorithms. Both inference processes led to recommendations with similar regret, but the SVD version was about 700 times faster, so we decided to use SVD for all other experiments.
    
    The state-of-the-art regarding bandit algorithms for PBM also includes PBM-UCB and PBM-PIE \cite{Lagree2016}, and PMED \cite{Komiyama2017}, which we do not consider in our experiments. While these three algorithms benefit from a theoretical analysis, our main focus here is the practical efficiency, both in terms of quality of the recommendation, and in terms of computational complexity. PBM-UCB and PBM-PIE have been shown \cite{Lagree2016} to lead to a higher regret than PBM-TS with which we compare \ouralgo{}. PMED is much slower than \ouralgo{} since it calls at each time-stamp a sub-routine with complexity $O(N^{4.5})$ and solves two optimization problems. 

\subsection{Results}
    We compare the previously presented algorithms on the basis of the \emph{regret} (see Equation \eqref{PseudoRegCum}), which is the sum, over  $T$ consecutive recommendations, of the difference between the expected reward of the best possible answer and of the answer of a given recommender system. The regret will be plotted with respect to $T$ on a log-scale basis. The best algorithm is the one with the lowest regret. The regret plots are bounded by the regret of the oracle (0) and the regret of a recommender system choosing the items uniformly at random. We average the results of each algorithm over 20 independent sequences of recommendations for the purely simulated data, and over the 8 queries (with 10 sequences of recommendations per query) for the behavioral data. 
    
    \subsubsection{\ouralgo{} hyper-parameters on purely simulated data}
    
    \ouralgo{} behavior is affected by two hyper-parameters: the width $c/\sqrt{t}$ of the Gaussian random-walk steps, and the number $m$ of Metropolis-Hastings iterations per recommendation. We show in Figure \ref{cpas} the impact of these hyper-parameters on purely simulated data. Note that the regret is the smallest for a wide range of hyper-parameters, which means that it is easy to tune \ouralgo{}'s hyper-parameters to obtain good recommendations. We get a high regret only when $c$ and $m$ are both too small (green curve): when the random-walk steps are too small, they are correlated and the Metropolis-Hasting algorithm requires more iterations to recommend relevant items. Overall, taking $c=100$ and $m=1$ is a good choice both in terms of regret and in terms of computation time, since the computation time of \ouralgo{} scales linearly with $m$.

\begin{figure}[tb]
    \centering
    \includegraphics[width=\linewidth]{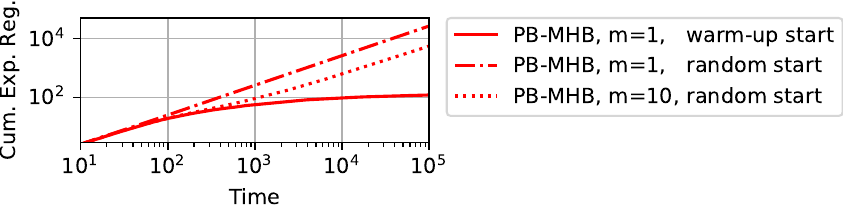}
    \caption{Cumulative regret w.r.t. time on purely simulated data with $\thetav$ close to real data. Impact of the use of the parameters from the previous time-stamp to warm-up the Metropolis-Hasting algorithm. The width parameter $c$ is fixed to 100.}
    \label{pasalea}
    \end{figure}

    \begin{figure*}[t]
    \centering
    \begin{subfigure}[b]{0.32\textwidth}
        \centering
        \includegraphics[width=\textwidth]{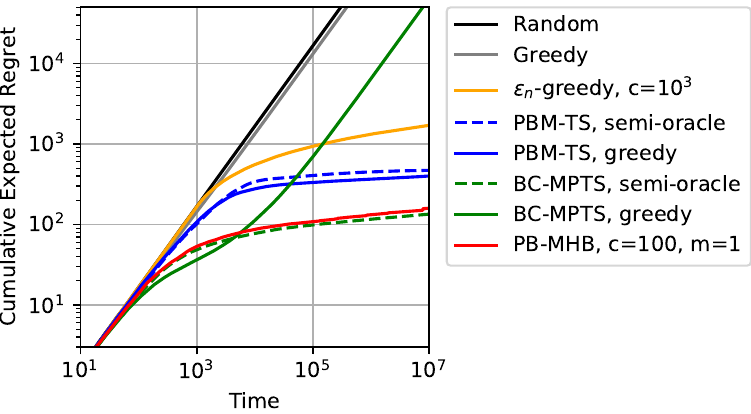}
        \caption{$\thetav$ close to 0}
        \label{fig:opponents_small}
    \end{subfigure}
    \hfill
    \begin{subfigure}[b]{0.32\textwidth}
        \centering
        \includegraphics[width=\textwidth]{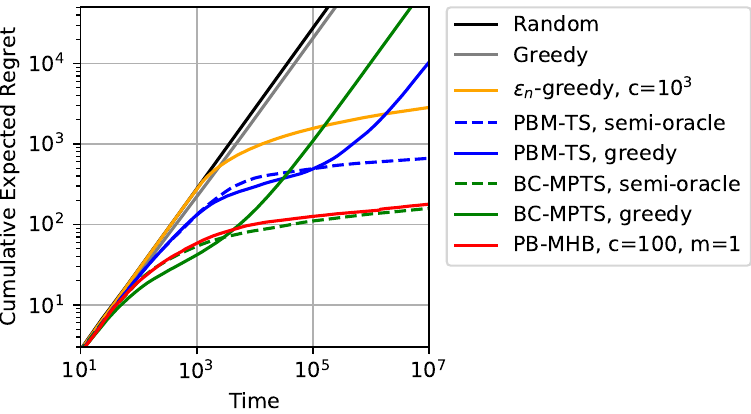}
        \caption{$\thetav$ close to real data}
        \label{fig:opponents_std}
    \end{subfigure}
    \hfill
    \begin{subfigure}[b]{0.32\textwidth}
        \centering
        \includegraphics[width=\textwidth]{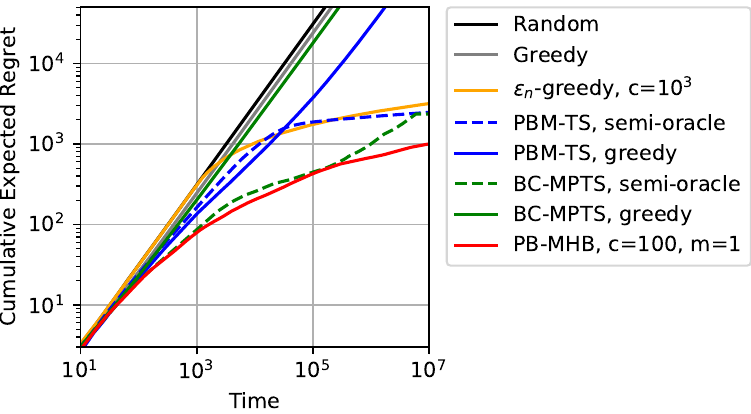}
        \caption{$\thetav$ close to 1}
        \label{fig:opponents_big}
    \end{subfigure}
    \caption{Cumulative regret w.r.t. time on purely simulated data for all competitors.}
    \label{opponents_Simu}
    \end{figure*}  

   \begin{figure}[tb]
    \centering
    \includegraphics[width=0.8\linewidth]{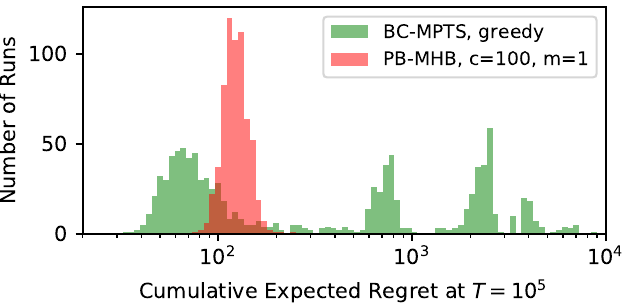}
    \caption{Distribution over $1,000$ runs of the cumulative regret at $T=10^5$ on purely simulated data with $\thetav$ close to real data.}
    \label{fig:compare_greedy}
    \end{figure}

     As \ouralgo{} initiates a Metropolis-Hastings run, it starts from the couple $(\tilde\thetav, \tilde\kappav)$ from the previous time-stamp. Figure \ref{pasalea} shows the impact of keeping the parameters from the previous time-stamp compared to a purely random start. Note that this warm-up start allows \ouralgo{} to have a small regret while only doing $m=1$ Metropolis-Hastings iterations per recommendation. Starting from a new randomly drawn set of parameters would require more than $m=10$ iterations to obtain the same result, meaning a computation budget more than 10 times higher. This behavior is explained by the gap between the uniform law (which is used to draw the starting set of parameters) and the targeted law (\emph{a posteriori} law of these parameters) which concentrates around its MAP. Even worse, this gap increases while getting more and more data since the \emph{a posteriori} law concentrates with the increase of data. As a consequence, the required value for $m$ increases along time when applying a standard Metropolis-Hasting initialisation, which explains why the dotted line diverges from the solid one around time-stamp 200.
   
    \subsubsection{Comparison on purely simulated data}

    Figure \ref{opponents_Simu} compares the regret obtained by \ouralgo{} and its competitors on purely simulated data.
    With the three settings, \ouralgo{} performs as well as the semi-oracle version of BC-MPTS, and has a smaller regret than any other algorithm. Remember that the semi-oracle version of BC-MPTS requires the knowledge of $\kappav$, while \ouralgo{} is learning it online.
    More specifically, when $\thetav$ is close to 1, \ouralgo{} even outperforms the semi-oracle version of BC-MPTS. With this setting, BC-MPTS does not explore enough, which was already observed by \cite{Lagree2016}.
    
    Also note that, as expected, the greedy versions of BC-MPTS and PBM-TS are too greedy: while they start with a regret similar to the regret of the semi-oracle versions, they quickly suffer a linear (averaged) cumulative regret. As can be observed in Figure~\ref{fig:compare_greedy}, the regret is the average between (i) "lucky" runs for which the algorithm focuses on the right configuration and enjoys a small regret, and (ii) "unlucky" runs for which the algorithm  focuses on the wrong configuration (it typically misses the right order for $\kappa$ values) and suffers a high regret. By exploring more, \ouralgo{} always suffers a greater regret than the one for "lucky guys", but never suffers a high regret. In average, this in-between strategy is the winning one.

    \begin{table}
    \centering
    \caption{Computation time for a sequence of $10^5$ recommendations vs. the purely simulated environment on an Intel Xeon E5-2450 CPU with 50 GB RAM. The algorithms are implemented in Python.}
    \label{tab:timing_simu}
    \begin{tabular}{llrr}
    \toprule
        \textbf{Algorithm} &&\textbf{Total time}& \textbf{Time per}\\
        \textbf{} &&\textbf{(min)}& \textbf{trial (ms)}\\
        \midrule
         $\varepsilon_n$-greedy& $c=10^3$ &0.5&$0.3$\\
         BC-MPTS& semi-oracle & 0.5&$0.3$\\
         & greedy & 1&0.6\\
         PBM-TS& semi-oracle & 7&4.2  \\
         & greedy & 8&4.8\\
         \ouralgo{}& $c=100$, $m=1$& 30&18\\
         & $c=100$, $m=10$& 4h11& 151\\
    \bottomrule
    \end{tabular}
    \end{table}
    
  All the algorithms (see Table \ref{tab:timing_simu}), require less than 20 ms per recommendation which remains affordable. As expected, for \ouralgo{}, doing $m=10$ Metropolis-Hastings iterations per recommendation increases the computation time by about 10 w.r.t. to $m=1$.

    \subsubsection{Comparison on behavioral data}

    
    
\begin{figure}[t]
    \centering
    \includegraphics[width=\linewidth]{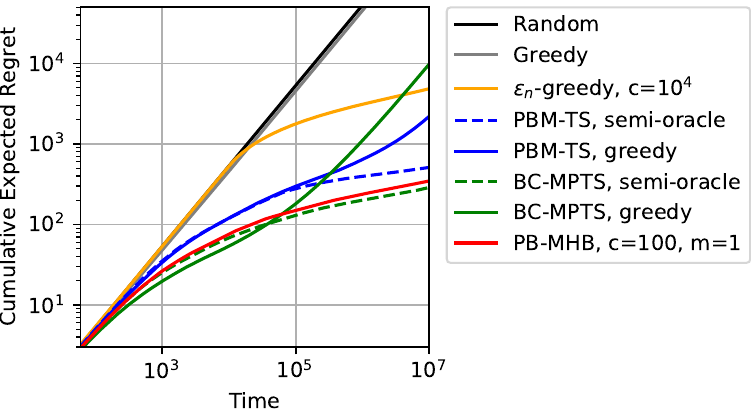}
    \caption{Cumulative regret w.r.t. time on behavioral data.}
    \label{opponents}
    \end{figure}

    The results with $(\kappav, \thetav)$ values extracted from the KDD Cup 2012 dataset are similar to the ones with purely simulated data (see Figure \ref{opponents}): 
    \ouralgo{} has the same regret as the semi-oracle version of BC-MPTS while \ouralgo{} does not require the knowledge of $\kappav$; other algorithms have a greater regret. Note that greedy versions of PBM-TS and BC-MPTS again suffer a linear regret due to a lack of exploration when $T$ is large enough.

    \section{Conclusion}
    \label{sec:conclu}
We presented a new bandit-based algorithm, \ouralgo{}, for online recommender systems in the PBM which uses a Thompson sampling framework to learn the $\kappav$ and $\thetav$ parameters of the PBM instead of assuming them given. Experiments on simulated and real datasets show that our method (i) suffers a smaller regret than its competitors with access to the same information, and (ii) suffers a similar regret as its competitors when  they use more prior information. These results are still empirical but we plan to formally prove them in future work. We also would like to improve our algorithm by further working both on the proposal law to draw candidates for the sampling part and on the update strategy for the target distribution which evolution is linked to the reward collection. The proposal is currently a truncated random walk. By managing it differently (with a logit transformation for instance) we could improve both the time and precision performance. On the other hand, with a better understanding of the evolution of the target distribution, we could also improve the sampling part.


\bibliographystyle{named}
\bibliography{theseCamille.bib}

\begin{thebibliography}{}

\bibitem[\protect\citeauthoryear{Agrawal and Goyal}{2017}]{Agrawal2017}
Shipra Agrawal and Navin Goyal.
\newblock Near-optimal regret bounds for thompson sampling.
\newblock {\em Jour. of the ACM, JACM}, 64(5):30:1--30:24, September 2017.

\bibitem[\protect\citeauthoryear{Aleksandr~Chuklin}{2015}]{Pyclick}
Maarten de~Rijke Aleksandr~Chuklin, Ilya~Markov.
\newblock Click models for web search.
\newblock In Kamalika Chaudhuri and Masashi Sugiyama, editors, {\em Click
  Models for Web Search}. Morgan and Claypool Publishers,
  http://clickmodels.weebly.com/the-book.html, 2015.

\bibitem[\protect\citeauthoryear{Auer \bgroup \em et al.\egroup
  }{2002}]{Auer2002}
Peter Auer, Nicol{\`o} Cesa-Bianchi, and Paul Fischer.
\newblock Finite-time analysis of the multiarmed bandit problem.
\newblock {\em Machine Learning}, 47(2):235--256, May 2002.

\bibitem[\protect\citeauthoryear{Chen \bgroup \em et al.\egroup
  }{2013}]{Chen2013}
Wei Chen, Yajun Wang, and Yang Yuan.
\newblock Combinatorial multi-armed bandit: General framework and applications.
\newblock In {\em proc. of the 30th Int. Conf. on Machine Learning}, ICML '13,
  2013.

\bibitem[\protect\citeauthoryear{Cheung \bgroup \em et al.\egroup
  }{2019}]{Cheung2019}
Wang~Chi Cheung, Vincent Tan, and Zixin Zhong.
\newblock A thompson sampling algorithm for cascading bandits.
\newblock In {\em proc. of the 22nd Int. Conf. on Artificial Intelligence and
  Statistics, AISTATS'19}, 2019.

\bibitem[\protect\citeauthoryear{Chuklin \bgroup \em et al.\egroup
  }{2015}]{Chuklin2015}
Aleksandr Chuklin, Ilya Markov, and Maarten de~Rijke.
\newblock {\em Click Models for Web Search}.
\newblock Morgan \& Claypool, 2015.

\bibitem[\protect\citeauthoryear{Combes \bgroup \em et al.\egroup
  }{2015}]{Combes2015}
Richard Combes, Stefan Magureanu, Alexandre Proutiere, and Cyrille Laroche.
\newblock Learning to rank: Regret lower bounds and efficient algorithms.
\newblock In {\em proc. of the 2015 ACM SIGMETRICS Int. Conf. on Measurement
  and Modeling of Computer Systems}, 2015.

\bibitem[\protect\citeauthoryear{Craswell \bgroup \em et al.\egroup
  }{2008}]{Craswell2008}
Nick Craswell, Onno Zoeter, Michael Taylor, and Bill Ramsey.
\newblock An experimental comparison of click position-bias models.
\newblock In {\em proc. of the 2008 Int. Conf. on Web Search and Data Mining},
  WSDM '08, 2008.

\bibitem[\protect\citeauthoryear{Katariya \bgroup \em et al.\egroup
  }{2016}]{Katariya2016}
Sumeet Katariya, Branislav Kveton, Csaba Szepesv\'{a}ri, and Zheng Wen.
\newblock Dcm bandits: Learning to rank with multiple clicks.
\newblock In {\em proc. of the 33rd Int. Conf. on Machine Learning}, ICML'16,
  2016.

\bibitem[\protect\citeauthoryear{Katariya \bgroup \em et al.\egroup
  }{2017a}]{Katariya2017a}
Sumeet Katariya, Branislav Kveton, Csaba Szepesvari, Claire Vernade, and Zheng
  Wen.
\newblock {Stochastic Rank-1 Bandits}.
\newblock In {\em proc. of the 20th Int. Conf. on Artificial Intelligence and
  Statistics, AISTATS'17}, 2017.

\bibitem[\protect\citeauthoryear{Katariya \bgroup \em et al.\egroup
  }{2017b}]{Katariya2017b}
Sumeet Katariya, Branislav Kveton, Csaba Szepesvári, Claire Vernade, and Zheng
  Wen.
\newblock Bernoulli rank-1 bandits for click feedback.
\newblock In {\em proc. of the 26th International Joint Conference on
  Artificial Intelligence, {IJCAI'17}}, 2017.

\bibitem[\protect\citeauthoryear{Kawale \bgroup \em et al.\egroup
  }{2015}]{Kawale2015}
Jaya Kawale, Hung~H Bui, Branislav Kveton, Long Tran-Thanh, and Sanjay Chawla.
\newblock Efficient thompson sampling for online matrix factorization
  recommendation.
\newblock In {\em Advances in Neural Information Processing Systems 28,
  NIPS'15}, 2015.

\bibitem[\protect\citeauthoryear{Komiyama \bgroup \em et al.\egroup
  }{2015}]{Komiyama2015}
Junpei Komiyama, Junya Honda, and Hiroshi Nakagawa.
\newblock Optimal regret analysis of thompson sampling in stochastic
  multi-armed bandit problem with multiple plays.
\newblock In {\em proc. of the 32nd Int. Conf. on Machine Learning}, ICML '15,
  2015.

\bibitem[\protect\citeauthoryear{Komiyama \bgroup \em et al.\egroup
  }{2017}]{Komiyama2017}
Junpei Komiyama, Junya Honda, and Akiko Takeda.
\newblock Position-based multiple-play bandit problem with unknown position
  bias.
\newblock In {\em Advances in Neural Information Processing Systems 30,
  NIPS'17}, 2017.

\bibitem[\protect\citeauthoryear{Kveton \bgroup \em et al.\egroup
  }{2015a}]{Kveton2015a}
Branislav Kveton, Csaba Szepesv\'{a}ri, Zheng Wen, and Azin Ashkan.
\newblock Cascading bandits: Learning to rank in the cascade model.
\newblock In {\em proc. of the 32nd Int. Conf. on Machine Learning}, ICML '15,
  2015.

\bibitem[\protect\citeauthoryear{Kveton \bgroup \em et al.\egroup
  }{2015b}]{Kveton2015b}
Branislav Kveton, Zheng Wen, Azin Ashkan, and Csaba Szepesv\'{a}ri.
\newblock Combinatorial cascading bandits.
\newblock In {\em Advances in Neural Information Processing Systems 28}, NIPS
  '15, 2015.

\bibitem[\protect\citeauthoryear{Lagr{\'e}e \bgroup \em et al.\egroup
  }{2016}]{Lagree2016}
Paul Lagr{\'e}e, Claire Vernade, and Olivier Capp{\'e}.
\newblock Multiple-play bandits in the position-based model.
\newblock In {\em Advances in Neural Information Processing Systems 30}, NIPS
  '16, 2016.

\bibitem[\protect\citeauthoryear{Li \bgroup \em et al.\egroup }{2016}]{Li2016}
Shuai Li, Baoxiang Wang, Shengyu Zhang, and Wei Chen.
\newblock Contextual combinatorial cascading bandits.
\newblock In {\em proc. of the 33rd Int. Conf. on Machine Learning, ICML'16},
  2016.

\bibitem[\protect\citeauthoryear{Neal}{1993}]{Neal1993}
Radford~M. Neal.
\newblock Probabilistic inference using markov chain monte carlo methods.
\newblock Technical Report CRG-TR-93-1, University of Zurich, Department of
  Informatics, 09 1993.

\bibitem[\protect\citeauthoryear{Richardson \bgroup \em et al.\egroup
  }{2007}]{Richardson2007}
Matthew Richardson, Ewa Dominowska, and Robert Ragno.
\newblock Predicting clicks: Estimating the click-through rate for new ads.
\newblock In {\em proc. of the 16th International World Wide Web Conference},
  WWW '07, 2007.

\bibitem[\protect\citeauthoryear{Thompson}{1933}]{Thompson1933}
William~R. Thompson.
\newblock On the likelihood that one unknown probability exceeds another in
  view of the evidence of two samples.
\newblock {\em Biometrika}, 25(3/4):285--294, 1933.

\bibitem[\protect\citeauthoryear{von Neumann}{1951}]{vonNeumann1951}
John von Neumann.
\newblock Various techniques used in connection with random digits.
\newblock In A.~S. Householder, G.~E. Forsythe, and H.~H. Germond, editors,
  {\em Monte Carlo Method}, volume~12 of {\em National Bureau of Standards
  Applied Mathematics Series}, chapter~13, pages 36--38. US Government Printing
  Office, Washington, DC, 1951.

\bibitem[\protect\citeauthoryear{Zong \bgroup \em et al.\egroup
  }{2016}]{Zong2016}
Shi Zong, Hao Ni, Kenny Sung, Nan~Rosemary Ke, Zheng Wen, and Branislav Kveton.
\newblock Cascading bandits for large-scale recommendation problems.
\newblock In {\em proc. of the 32nd Conference on Uncertainty in Artificial
  Intelligence}, UAI '16, 2016.

\end{thebibliography}

\end{document}